\documentclass[letterpaper, 10 pt, conference]{ieeeconf}

\IEEEoverridecommandlockouts 
\usepackage{amsmath}                      
\usepackage{amssymb}                      
\usepackage{algorithm}                    
\usepackage{algpseudocode}
\usepackage[T1]{fontenc}                  
\usepackage{graphicx}
\usepackage{subcaption}                  
\usepackage[urlcolor=blue,colorlinks=true]{hyperref} 
\usepackage[utf8]{inputenc}               
\usepackage{soul}                         
\usepackage[usenames,dvipsnames]{xcolor}  
\usepackage{multirow}                     
\usepackage{makecell}					  

\newcommand{\dof}{{\sc dof}}

\newcommand{\cspace}{\ensuremath{\mathcal{C}_{space}}}
\newcommand{\cspaces}{\ensuremath{\mathcal{C}_{spaces}}}

\newcommand{\env}{\mathcal{E}}
\newcommand{\robots}{\mathcal{R}}
\newcommand{\query}{\mathcal{Q}}
\newcommand{\paths}{\mathcal{P}}
\newcommand{\conflict}{\mathcal{C}}

\newcommand{\algorithmicinput}{\textbf{Input:}}
\newcommand{\algorithmicoutput}{\textbf{Output:}}
\newcommand{\INPUT}{\item[\algorithmicinput]}
\newcommand{\OUTPUT}{\item[\algorithmicoutput]}

\algdef{SE}[DOWHILE]{Do}{doWhile}{\algorithmicdo}[1]{\algorithmicwhile\ #1}%

\begin{document}
	
	\title{\LARGE \bf
		Adaptive Robot Coordination: A Subproblem-based Approach for Hybrid Multi-Robot Motion Planning }
	
	\author{Irving Solis$^{1}$, James Motes$^{2}$, Mike Qin$^{2}$, Marco Morales$^{2}$, and
		Nancy M. Amato$^{2}$
		\thanks{$^{1}$Irving Solis is with the Texas A\&M
			University Department of Computer Science and Engineering, College Station, TX,
			77840, USA. \{irvingsolis89\}@tamu.edu.}%
		\thanks{$^{2}$James Motes, Mike Qin, Marco Morales and Nancy M. Amato are with the University of Illinois Urbana-Champaign
			Department of Computer Science, Urbana, IL, 61801, USA. \{jmotes2, yudiqin2, moralesa, namato\}@illinois.edu.}%
	}
	
	\maketitle
	
	\thispagestyle{empty} 
	\pagestyle{plain}     
	
	
	\begin{abstract}


  This work presents Adaptive Robot Coordination (ARC), a novel hybrid framework for multi-robot motion planning (MRMP) that employs local subproblems to resolve inter-robot conflicts. ARC creates subproblems centered around conflicts, and the solutions represent the robot motions required to resolve these conflicts. The use of subproblems enables an inexpensive hybrid exploration of the multi-robot planning space. ARC leverages the hybrid exploration by dynamically adjusting the coupling and decoupling of the multi-robot planning space. This allows ARC to adapt the levels of coordination efficiently by planning in decoupled spaces, where robots can operate independently, and in coupled spaces where coordination is essential. ARC is probabilistically complete,  can be used for any robot, and produces efficient cost solutions in reduced planning times. Through extensive evaluation across representative scenarios with different robots requiring various levels of coordination, ARC demonstrates its ability to provide simultaneous scalability and precise coordination. ARC is the only method capable of solving all the scenarios and is competitive with coupled, decoupled, and hybrid baselines.

	\end{abstract}
		
	\section{Introduction}

Multi-robot systems (MRS) have gained significant prominence in various applications, including payload transportation and manufacturing, due to their ability to enhance productivity and reduce operational costs. 
Multi-robot Motion Planning (MRMP) is the problem of determining feasible paths for a given MRS. 
The complexity of MRMP typically arises from the quantity of robots in play and the degree of coordination required to tackle the problem. In this paper, our attention is directed towards the challenges associated with motion planning when both of these aspects come into play.
Indeed, these aspects represent a trade-off illustrated by the differences between coupled and decoupled MRMP approaches. 
As a result, the latest research has focused on studying hybrid approaches, which combine decoupled and coupled behaviors to obtain scalability and enhanced coordination. Hybrid techniques generally first behave decoupled to obtain individual paths and then behave coupled to resolve conflicts between robots.

In this work, we explore a novel way to resolve conflicts. Instead of an immediate transition to a coupled search, we first define local subproblems around conflicts, the solutions to which correspond to appropriate robot motions needed to resolve the conflict. Local subproblems can adjust themselves to only consider the relevant robots and the needed portion of their corresponding composite space. Subproblems allow the discovery of new states and a flexible (de)coupled search instead of a straight coupled search in the existing state space representation.





    \begin{figure}[h!]
		\centering
		\begin{subfigure}[b]{0.3\linewidth}
			\includegraphics[width=\linewidth]{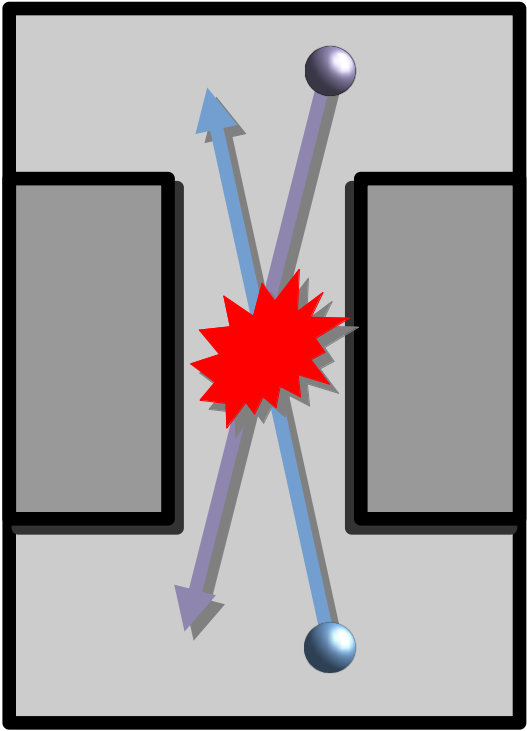}
			\caption{}
		\end{subfigure}
		\begin{subfigure}[b]{0.3\linewidth}
			\includegraphics[width=\linewidth]{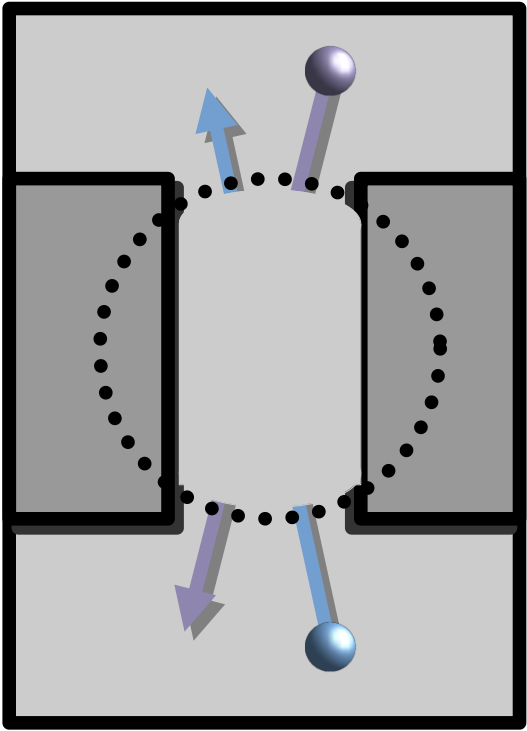}
			\caption{}
		\end{subfigure}
		\begin{subfigure}[b]{0.3\linewidth}
			\includegraphics[width=\linewidth]{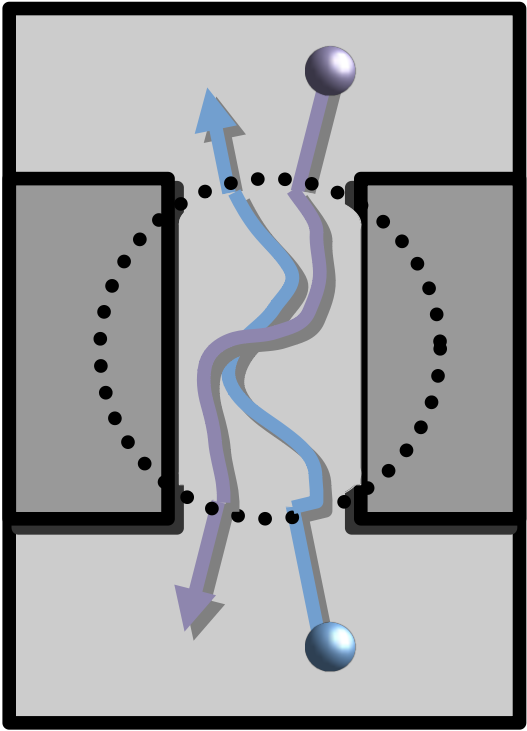}
			\caption{}
		\end{subfigure}
		\caption{{\small{{A simplified overview of our method: a) Detect a conflict between two robots' paths. b) Define a local subproblem around the conflict. c) Use the subproblem's solution to modify the original paths and resolve the conflict. }}      
        \vspace{-1mm}}}
		\label{fig:overview}
	\end{figure}



In this paper, we present Adaptive Robot Coordination (ARC), a MRMP hybrid framework that models conflict resolutions as local subproblems and uses their solutions to obtain feasible paths. 
ARC first constructs decoupled representations to obtain individual paths and dynamically generates coupled representations in response to conflicts through the formulation of subproblems. ARC provides probabilistic completeness by employing efficient heuristics for rapid subproblem adaptation, determining the robots involved and the composite space region required for achieving coordination. 
As a result, subproblems enable a feasible and non-expensive composite space search. 
Our method establishes a novel hybrid way to search the planning space by reasoning about different levels of (de)coupled spaces, enabling adaptive levels of robot coordination.

We demonstrate the ability of ARC to adapt robot coordination efficiently through a comprehensive set of experiments involving mobile, planar manipulator, and 3D manipulator robots in scenarios that demand low, high, and varying levels of coordination. We compare it against decoupled, hybrid, and composite baselines. Notably, ARC stands out as the sole method that successfully solves all the scenarios. In the low coordination scenario, ARC performs on par with the decoupled baseline. Similarly, in high coordination scenarios, ARC exhibits behavior similar to the composite baseline. In scenarios that require different levels of coordination, ARC outperforms all the baselines.

	In summary, our contributions are:
	\begin{itemize}
        \item A novel subproblem-based approach to resolve robot conflicts in MRMP problems.
        \item A MRMP method that exploits this framework to adapt robot coordination by transitioning different levels of (de)coupled spaces to find feasible solutions to problems with many robots requiring high levels of coordination. 
        \item Experimental evaluation  of this proposed method with up to 32 mobile robots and 16 manipulator robots. Results show efficient robot coordination while maintaining scalabilty. Our method exhibits improved performance over existing methods and is the only method able to solve all the scenarios.
        
        \end{itemize}

\section{Preliminaries and Related Work}
	
In this section, we first provide insight into single-robot and multi-robot motion planning. We then review relevant MRMP algorithms. We look closely  at how these methods explore multi-robot state space, coupled and decoupled. We then discuss how our method fits and differs from the current literature. 
	
\subsection{Motion Planning}
	
Motion Planning is the problem of finding a valid path for a given robot, from a start to a goal pose.
Solutions involve exploring and finding continuous paths in the \textit{configuration space} (\cspace)~\cite{lw-apcfpapo-79}, the set of all the robot's configurations. 
A configuration encodes the values of the robot's degrees-of-freedom (\dof), which are the parameters that define the robot state (i.e., position, orientation, joint angles, velocity). 
As the number of {\dof} increases, representing the {\cspace} becomes intractable~\cite{c-crmp-88,r-cpmpg-79}.
Sampling-based algorithms like the Probabilistic Roadmap Method (PRM)~\cite{kslo-prpp-96} were developed as approximate solutions which trade completeness for probabilistic completeness.
PRMs capture the \cspace \hspace{1mm}connectivity by sampling graphs, also known as roadmaps, whose vertices represent valid robot states and edges valid transitions between states. Paths are obtained by querying these roadmaps. Finding a feasible solution can become challenging depending upon the connectivity, such as in environments with narrow passages ~\cite{hklms-fnp-98}. 

	
\subsection{Multi-robot Motion Planning}

Multi-robot motion planning is the problem of finding a set of feasible paths for a given set of robots, each with start and goal poses. Solutions require exploring the composite configuration space, the state space that encodes all the robots' individual configuration spaces.  The composite space can be explored in a coupled and decoupled manner. The former consists of exploring the composite space directly, for example, sampling a roadmap to find a path from a start composite state to a goal composite state. The latter decomposes the composite space into individual configuration spaces, for example, sampling individual roadmaps on each robot's configuration space to find its corresponding path. Decoupled paths must be adjusted later to avoid conflicts. The following subsections discuss how coupled and decoupled methods address MRMP in further detail.

\subsection{Coupled methods}

Coupled approaches explore the composite space directly to find a path from a start to a goal composite configuration. Start and goal composite configurations encode all the robots' starts and goals. As the composite configuration space is intractable, these techniques approximate it by constructing a representation of the space. Sampling-based approaches such as PRM~\cite{kslo-prpp-96} and RRT~\cite{l-rrtntpp-1998} can be adapted to explore the composite space and find multi-robot solutions ~\cite{sl-uppccdpmrs-2002,cp-prrtmrs-02,fks-rwr-06,mnvp-marsbcp-13,ssh-faniaehdrfeoirimrmp-16,si-mrmpbic-06}. Exploring the composite space directly provides probabilistic completeness and high levels of coordination, allowing to solve complex problems such as a pair of mobile robots crossing an inlet or planning for two tangled robotic arms. However, these methods are only suitable for fewer robots, as exploring the composite space directly becomes strenuous when the number of robots increases.

Multi-robot Pathfinding (MRPF) techniques ~\cite{p-hissfcps-84,so-cppfmr-98} have been adapted for MRMP problem-solving by exploring a composite state space derived from the cartesian product of individual state spaces.  However, their effectiveness relies heavily on the quality of individual representations and may falter if they lack essential coordination states.

\subsection{Decoupled methods}

Decoupled methods decompose the space into individual configuration spaces, and each path is found by exploring its corresponding configuration space. Still, each configuration space is also intractable, so a representation must be constructed on each one. Sampling-based algorithms can be used to construct each representation. Due to decoupled representations being built in isolation, single-robot search algorithms can not be directly used to obtain a feasible multi-robot solution. Instead, Multi-robot Pathfinding (MRPF) algorithms are adapted to query the decoupled representations to address this issue. 

MRPF techniques have also been adapted to query decoupled representations. Dynamic obstacle-based techniques plan feasible paths by computing an individual robot’s path and considering the other robots as dynamic obstacles. Priority-based solvers assign priorities in which to plan in order to reduce conflicts. Paths are then adjusted to avoid the remaining conflicts.    

\subsection{Hybrid methods} 

Hybrid approaches combine the decoupled and coupled behaviors to obtain scalability and enhanced coordination. Efficient MRPF  ~\cite{wc-sefmpp-15,ssfs-cbsfomap-15} hybrid techniques have been proposed to resolve discrete and more relaxed problems. 
They start by computing individual paths in each robot configuration space. When robots are found to collide (conflict), ~\cite{wc-sefmpp-15} M* expands the search space's dimensionality to identify the coupled actions needed to resolve the conflict, while ~\cite{ssfs-cbsfomap-15} CBS imposes constraints on the individual configuration spaces to facilitate the re-planning of collision-free paths.

Hybrid MRMP planners have also been proposed ~\cite{smsm-romrmpucbs-21,cukk-oabsmamp-19}. These planners extend the CBS framework to solve multi-robot sampling-based and state-lattice motion planning problems. These techniques adapt the CBS hybrid framework to explore a planning space derived from the individual robot state space representations.  Notably, these representations are constructed decoupled, making it uncertain if they contain essential coordination states to obtain a solution. Consequently, these techniques will likely struggle in scenarios that demand precise coordination, where exploring higher-dimensional spaces is critical.

This contrasts our method, which, by introducing local subproblems, can effectively conduct both coupled and decoupled explorations at different levels of state space compositions, facilitating the discovery of new states essential for robot coordination.

	\section{Problem Definition}
	\label{section:problem-definition}
	
	In this section we define the sampling-based MRMP problem and essential concepts.

	A multi-robot motion planning (MRMP) problem is defined by $(\mathcal{E},\mathcal{R},\mathcal{Q})$, where $\mathcal{E}$ represents the environment, $\mathcal{R}$ is a set of robots, and $\mathcal{Q}$ is a set of queries  consisting of a start and goal position for each robot.
	The configuration space for each robot $r_i\in\mathcal{R}$ is denoted $\mathcal{C}_i$.
	The composite configuration space of the entire system $\mathcal{C}_\texttt{composite}$ is the cartesian product of all individual robot configuration spaces.
	
	A MRMP method must find a path through $\mathcal{C}_\texttt{composite}$ which transitions each robot $r_i\in\mathcal{R}$ from its start state $q^\texttt{start}_i\in\mathcal{Q}$ to its goal state $q^\texttt{goal}_i\in\mathcal{Q}$.
	Decoupled planners decompose $\mathcal{C}_\texttt{composite}$ into individual robot configuration spaces, and plan initial paths for the robots.
	This requires the detection and resolution of \textit{conflicts} between individual paths to find a valid solution.
	A \textit{conflict} is defined as the event when two robots $r_i$ and $r_j$interfere with each other at timestep $t$ while traversing their paths. 
	It is denoted $<c_i,c_j,t>$ where $t$ is the timestep when the conflict occurred, and $c_i$ and $c_j$ are corresponding configurations of the conflicting robots. 

	A MRMP solution is valid if all robot paths are conflict-free and transition the robots from their start states to their goal states.

\section{Method}
\label{method}

In this section, we present the Adaptive Robot Coordination (ARC) approach to the multi-robot motion planning (MRMP) problem.
We first provide an overview of the method and how subproblems are used to resolve conflicts. Then, we detail the creation and adaptation of subproblems.
Finally, we discuss the theoretical properties of the approach.

\subsection{Overview}
\label{method:overview}
ARC is a hybrid MRMP method that employs subproblems to efficiently address conflicts by exploring relevant sections of the planning space.
Given a MRMP problem instance $(\env,\robots,\query)$, ARC (Alg.\ref{alg:ARC}) begins by obtaining the initial set of paths. Paths are found by solving each robot's individual motion planning problem $(\env,\robots_i=\{r_i\},\query=\{q_i\})$  using probabilistically complete sampling-based techniques.
These paths are represented by a sequence of configurations approximating continuous motion, and $p_i(t)$ is the configuration along $p_i$ at timestep $t$.

Because paths are computed in isolation, we check them for conflicts (Alg. \ref{alg:ARC}: lines \ref{alg:first_conflict},\ref{alg:find_additional_conflict}).
Conflicts between paths $p_i,p_j$ are used to create local subproblems $(\env',\robots'= r_i\cup r_j,\query')$ (Alg. \ref{alg:ARC}: line \ref{alg:create_subproblem}).
$\query'$ is defined by selecting points along $p_i,p_j$ sufficiently before and after the conflict timestep.
$\query'$ is used to define a local region $\env'$ for the subproblem.
Sampling-based MRMP solvers are used to find the subproblem solution, each with specific termination criteria. Local paths from the subproblem solution resolve the conflict and are connected to the rest of the initial paths. This process is repeated until resolving all conflicts to obtain a feasible MRMP solution. 

\begin{figure}[h!]
		\centering
		\begin{subfigure}[b]{0.49\linewidth}
			\includegraphics[width=\linewidth]{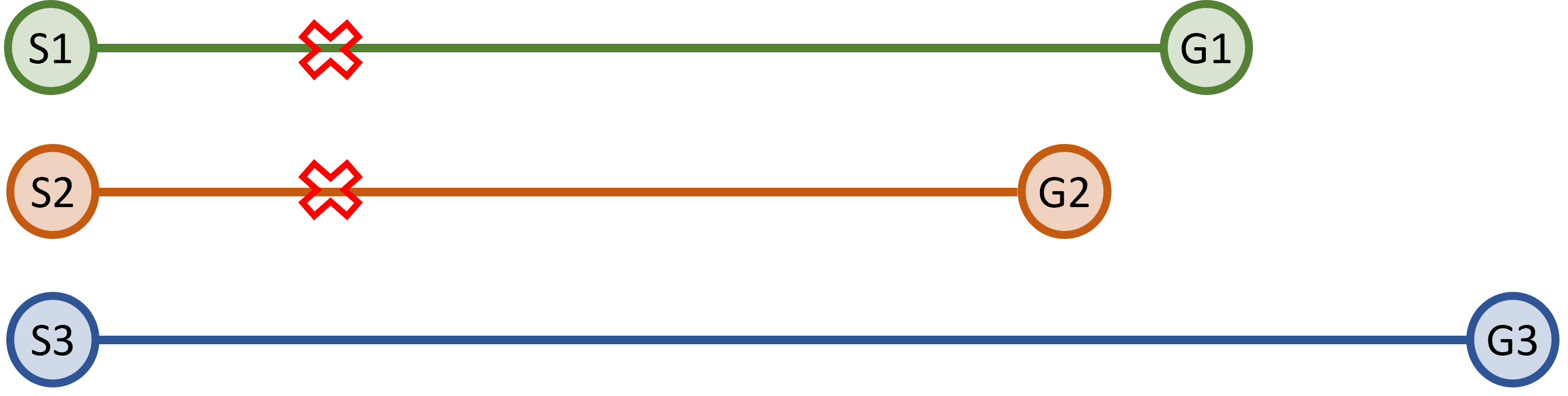}
			\caption{}
		\end{subfigure}
		\begin{subfigure}[b]{0.49\linewidth}
			\includegraphics[width=\linewidth]{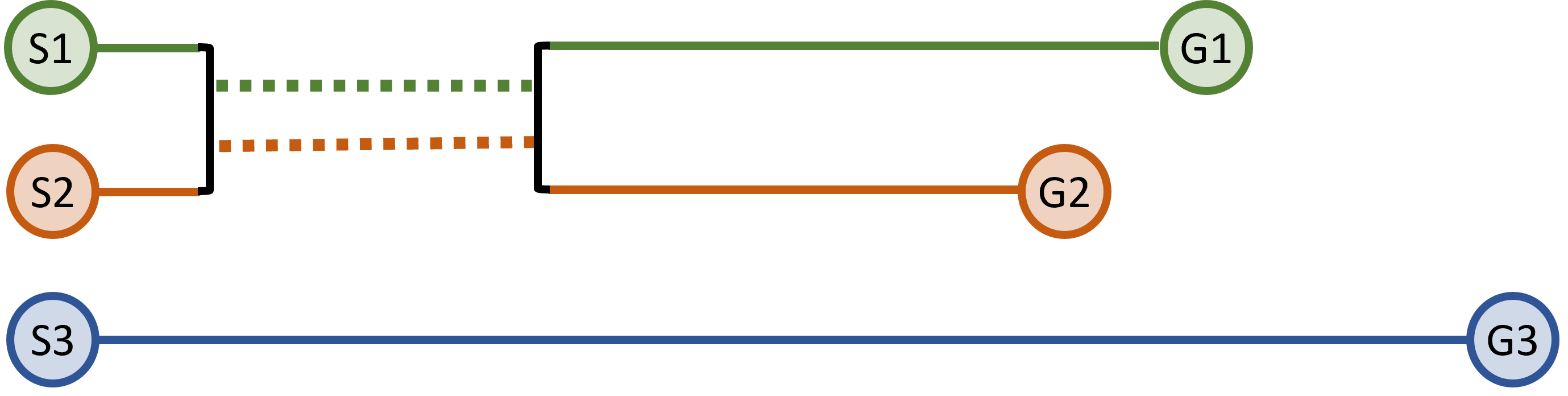}
			\caption{}
		\end{subfigure}
		\begin{subfigure}[b]{0.49\linewidth}
			\includegraphics[width=\linewidth]{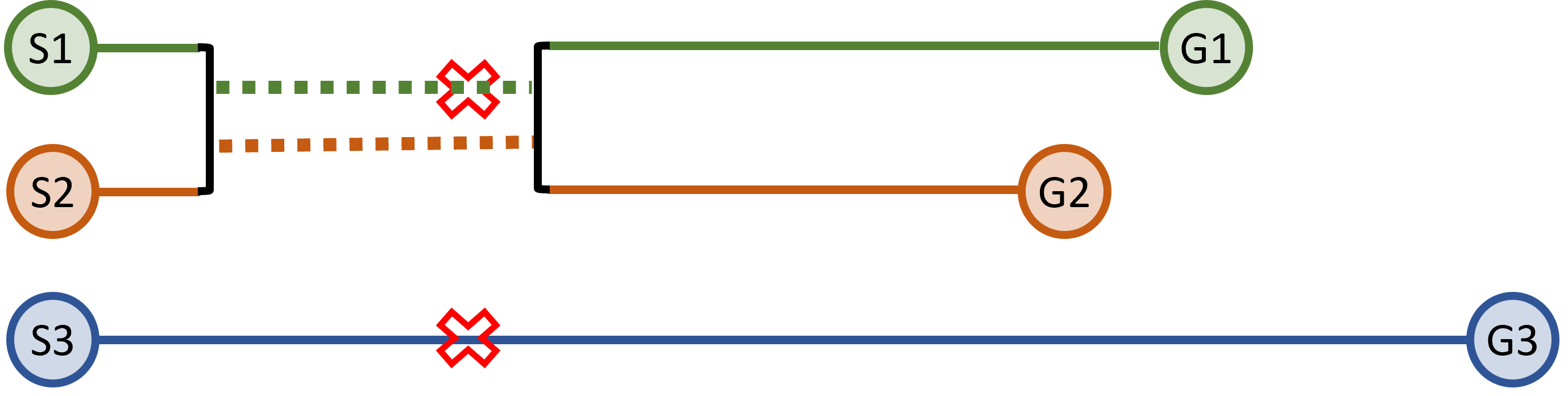}
			\caption{}
		\end{subfigure}
        \begin{subfigure}[b]{0.49\linewidth}
			\includegraphics[width=\linewidth]{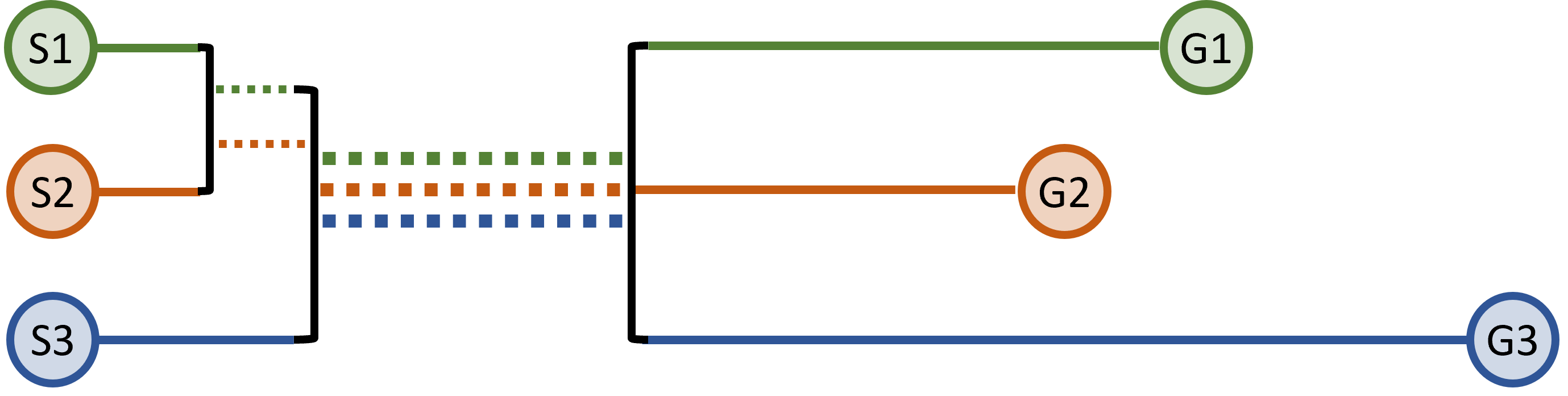}
			\caption{}
		\end{subfigure}
		\caption{{\small{"A three-robot MRMP problem is illustrated with path timelines indicating conflict times. Dashed lines represent path modifications to resolve conflicts. a) The initial conflict involves robots 1 and 2, resolved by defining a subproblem. b) The subproblem's solution resolves the conflict. c) A second conflict emerges as Robot 3 conflicts with the subproblem's solution for robots 1-2. d) A new subproblem, involving robots 1, 2, and 3, is defined and solved." \vspace{-2mm}}}}
		\label{fig:merging_subproblem}
\end{figure}

\begin{figure}[h!]
		\centering
		\begin{subfigure}[b]{0.3\linewidth}
			\includegraphics[width=\linewidth]{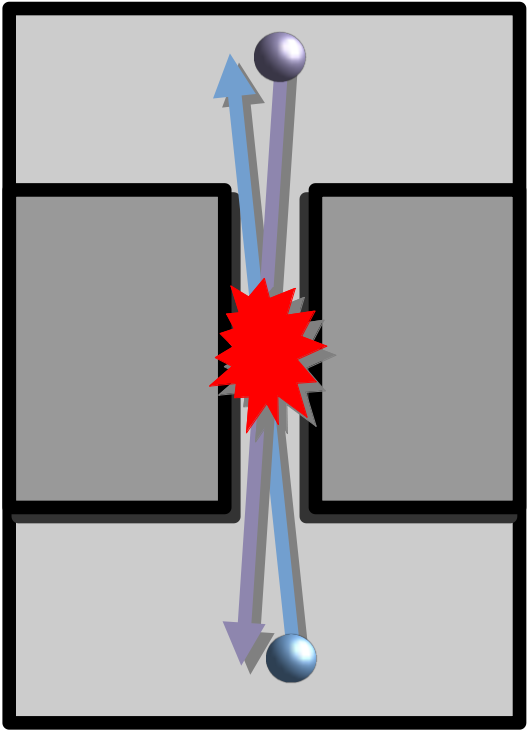}
			\caption{}
		\end{subfigure}
		\begin{subfigure}[b]{0.3\linewidth}
			\includegraphics[width=\linewidth]{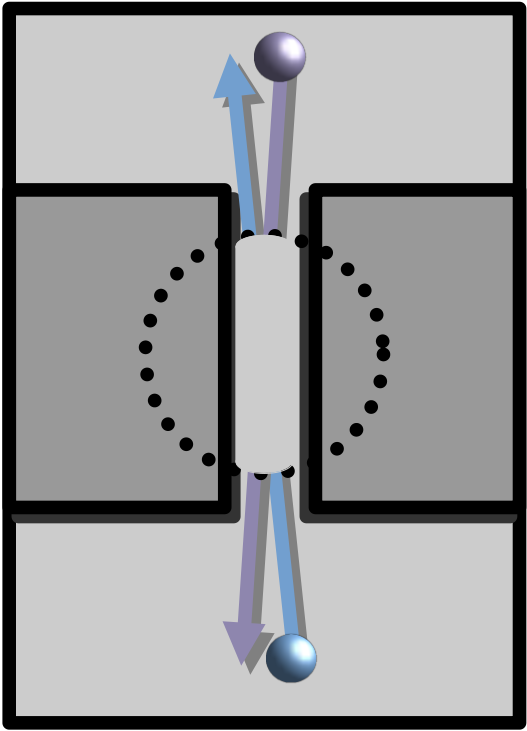}
			\caption{}
		\end{subfigure}
		\begin{subfigure}[b]{0.3\linewidth}
    		\includegraphics[width=\linewidth]{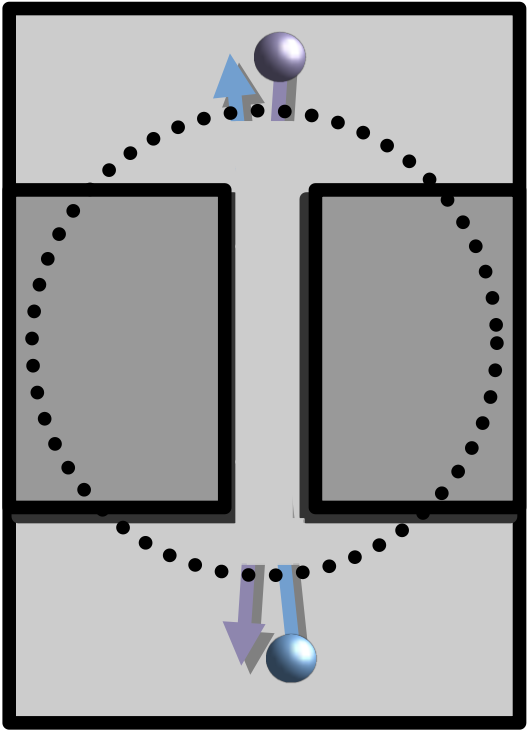}
			\caption{}
		\end{subfigure}
		\caption{{\small{a) A conflict between two paths inside a narrow passage. b) Initial subproblem where no solution exists. c) Expanding subproblem to find a feasible solution.\vspace{-1mm}}}}
		\label{fig:expand_subproblem}
	\end{figure}

If the probabilistically complete, sampling-based motion planning approach fails to find a solution, the local subproblem  $(\env',\robots',\query')$ is adapted by pushing $\query'$, further from the conflict on $p_i,p_j$.
$\env'$ is expanded accordingly (Alg. \ref{alg:solve_subproblem}: line \ref{alg:expand_subproblem}, Fig. \ref{fig:expand_subproblem} ) and this enlarged subproblem is attempeted to be solved. 
 
The local solution $\paths'$ to $(\env',\robots',\query')$ resolves the conflict in $p_i,p_j$.
In the case $p'$ conflicts with another subproblem solution, we introduce a new local subproblem that accounts for all conflicting robots.  (Alg. \ref{alg:ARC}: line \ref{alg:create_subproblem}, Fig. \ref{fig:merging_subproblem}). ) We discuss this in more detail in the following subsections. 

\begin{figure}[h!]
		\centering
    		  \includegraphics[width=1\linewidth]{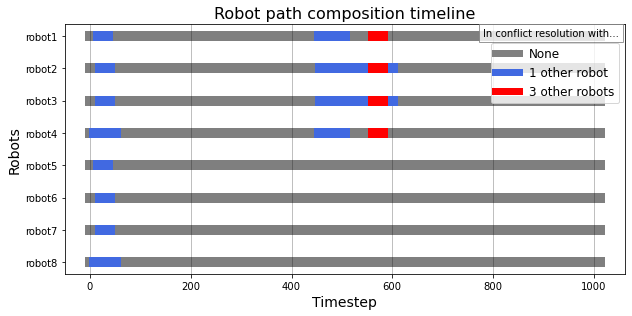}
		\caption{\small Timeline for each robot's path with colors representing segments planned at different state space compositions.  This is a solution of the problem shown in Fig. \ref{fig:AdaptiveCoordination-a}. 
  Paths are primarily computed within individual state spaces but transition to composite state spaces as higher levels of coordination become necessary for conflict resolution.
In the initial stages, ARC transitions to higher-level state spaces to address conflicts between pairs of robots of the same color. Around timestep 600, ARC transitions to an even more elevated state space to resolve a conflict involving four differently colored robots at the center.
  }
		\label{fig:path-composition}
	\end{figure}

The final solution results in a set of paths $\mathcal{P}$ in different compositions of robot state spaces (\ref{fig:path-composition}).
Each $p_i\in \paths$ has a start and end timestep $p_i.t_\texttt{start},p_i.t_\texttt{end}$ and configuration $p(p_i.t_\texttt{start}),p(p_i.t_\texttt{end})$.
There exists a $p_{R_i},p_{R_i}'\in P$ such that $R_i=\{r_i\}$ for all $r_i\in\robots$ and $p_{R_i}(0)=q_i.\texttt{start}$ and $p_{R_i}'(t_\texttt{final})=q_i.\texttt{goal}$.
If $r_i$ is never found to be in conflict with another robot, then $p_{R_i}(0)=p_{R_i}'$.

At every change of robot state space composition from a set of paths $P_\texttt{pre}$ to a set of paths $P_\texttt{post}$, the end configuration of the pre-transition paths $\prod_{p_i\in P_\texttt{pre}}p_i(p_i.t_\texttt{end})$ is equivalent to the start configuration of the post-transition paths $\prod_{p_i\in P_\texttt{post}}p_i(p_i.t_\texttt{start})$.
Other than timesteps where a robot is part of a state space composition change, each robot $r_i$ should exist in exactly one path at every timestep. An example of this process is illustrated in Fig. \ref{fig:path-composition}.




\begin{algorithm}
		\caption{Adaptive Robot Coordination (ARC)}\label{alg:ARC}
		\label{alg:overview}		
		\begin{algorithmic}[1]
			\INPUT A MRMP problem with an environment $\env$, a set of robots $\robots$, a set of queries $\query$.
			\OUTPUT A set of valid paths $\paths$.
			\State $\paths\leftarrow\emptyset$ \label{alg:empty_paths_1}
			\For {each robot $r_i \in \robots$ } \label{alg:initial_paths}
			\State $p_i\leftarrow$\texttt{MotionPlanning}($\env,\{r_i\},\{q_i\}$)
			\State $\paths\leftarrow \paths\cup\{p_i\}$
			\EndFor \label{alg:empty_paths_2}
			\State {$\conflict$ = \texttt{FindConflict($\paths$)}} \label{alg:first_conflict}
			\While {$C$ $\neq \emptyset$}
                \State \texttt{$\env'$,$\robots'$,$\query'$ = CreateSubProblem($\conflict,\paths,\env$)} \label{alg:create_subproblem}
                \State \texttt{$\paths'$ = SolveSubProblem($\env', \robots', \query'$)}
                \If {$\paths' != \emptyset$} \Comment{conflict resolved}
                    \State \texttt{UpdateSolution($\paths,\paths'$)} 
                    \State {$\conflict$ = \texttt{FindConflict($\paths$)}}\label{alg:find_additional_conflict}
                \Else 
                    \State $\paths = C \leftarrow \emptyset$ \Comment{if conflict not resolved,}
                \EndIf \Comment{return empty solution}
			\EndWhile
			\State\Return $\paths$\label{alg:return_solution}			

		\end{algorithmic} 
	\end{algorithm}

	\begin{algorithm}
		\caption{SolveSubProblem}\label{alg:solve_subproblem}
		\begin{algorithmic}[1]
			\INPUT A subproblem with an environment $\env'$, a set of robots $\robots$, a set of queries $\query'$.
                A set of MRMP solvers $\mathcal{S}$.
			\OUTPUT A set of valid local paths $\paths'$.
            \State $\paths'\leftarrow\emptyset$
            \While{$\paths'== \emptyset$}  \label{alg:while}		
                \For {$s$ \textbf{in} $\mathcal{S}$ } \label{alg:solvers_subproblem}       
        			\State {$\paths'\leftarrow$\texttt{SolveMRMP($s,\env',\robots',\query'$)}} \label{alg:solve_local_subproblem}
        			\If{$\paths' \neq \emptyset$} \Comment{subproblem solved}
                        \State{}
        			    \Return {$\paths'$}
        			\EndIf
                \EndFor
                \If {$\env' \neq \env$ \textbf{or} $\query' \neq \query$}
                    \State {\texttt{AdaptSubProblem($\env',\robots',\query'$)}}\label{alg:expand_subproblem}
                    
                \Else 
                    \State {}
                    \Return $\emptyset$  \label{alg:fail_subproblem} \Comment{if subproblem not solved,}
                \EndIf \Comment{return empty local solution}
            \EndWhile
			
		\end{algorithmic}
	\end{algorithm}

\subsection{Subproblem Creation and Adaptation}
\label{method:subproblem-creation}
From another perspective, conflict resolution involves the required robot motions to evade the conflict. Consequently, this can be represented as a local subproblem, with the solution being the necessary motions.
Given a set of paths $P$, and a conflict $(c_i,c_j,t)$ between $p_{R_i},p_{R_j}\in P$, we define a local subproblem $(\env',\robots',\query')$ around the conflict  (Alg. \ref{alg:ARC}: line \ref{alg:create_subproblem}).
$\robots'=R_i\cup R_j$ merges the involved robots.
The local query $\query'$ comprises the local start and goal configurations for each robot and is obtained by taking the corresponding configurations located in a time window before and after the conflict timestep $t-\texttt{window},t+\texttt{window}$, where the time window consists of an initial number of timesteps. 
The local region $\env'$ is defined by a $\cspace$ boundary encapsulating $\query'$.
This allows the planning methods to focus the $\robots'$ composite space search on a local region around the conflict.

In case a solution cannot be found for $(\env',\robots',\query')$, the local problem is adapted by expanding $\query'$ and $\env'$(Alg. \ref{alg:solve_subproblem}: line \ref{alg:expand_subproblem}). 
This adaptation involves the continuous advancement of the query points along their respective paths, which in turn expands the local environment.
This expansion process is repeated until a solution is found or $\env'=\env$ and $\query'=\query$ at which point the method terminates if no solution is found (Alg. \ref{alg:solve_subproblem}: line \ref{alg:fail_subproblem}).

If additional robots need to be incorporated to resolve the current conflict, ARC has the capability to adapt the set of $\robots'$. 
In scenarios involving multiple robots interacting in confined spaces, there are instances where resolving one conflict invalidates a prior conflict resolution.
ARC can identify such occurrences and adapt $\robots'$ to account for all the involved robots, ensuring a feasible resolution for all of them.

\subsection{Subproblem Planning}
\label{method:subproblem-planning}

While subproblems allow focusing computational effort on conflict resolution, not all conflicts occur equal, so ARC can also adapt the complexity of the method used to solve subproblems to the level of coordination required.
We do this through the use of a hierarchy of multi-robot motion planning (MRMP) methods (Alg. \ref{alg:solve_subproblem}, line \ref{alg:solvers_subproblem}), each enabling higher levels of coordination at the expensive computational effort.
The framework can be customized to obtain different behaviors by using different methods in the hierarchy.
We use the following hierarchy consisting of basic MRMP methods:

\begin{itemize}
\item Prioritized Query (no sampling)
\item Decoupled PRM (sampling individual robot $\cspaces$)
\item Composite PRM (sampling composite $\cspace$)
\end{itemize}

The sequence allows us to first attempt to resolve the conflict using our existing representations of the decoupled $\cspaces$.
This often involves one or more robots waiting for the others to pass the conflict region.

We follow this up with Decoupled PRM, which expands the roadmap representations of the decoupled $\cspaces$.
When waiting is insufficient, adding additional configurations that robots can use to move out of the way often provides a resolution.
These decoupled approaches failing usually indicate that the conflict requires more coordination. Thus, we use Composite PRM to build and search the composite space directly. 
If the final layer of the hierarchy fails, the local subproblem is expanded.

As both sampling methods are probabilistically complete, it is necessary to set termination criteria so that the methods can terminate and move on to the next approach.
While this can be adjusted to toggle the effort spent by each method, we have found for the decoupled approaches, it is better to fail fast and escalate the level of coordination used or the scope of the subproblem. In the case of the coupled approach Composite PRM, we leverage the metrics in ~\cite{mpa-maecm-06} to develop an adaptive termination criteria which assess the progress on exploring the composite space to set a limit on the exploration of new states. This criteria is an essential component of our method as it sets the appropriate effort on exploring the composite space, which is more expensive. 
In Section V, we explore scenarios in which each local method provides the desired level of coordination to resolve different types of conflicts. 

\subsection{Theoretical Properties}
\label{method:theoretical-properties}
When ARC can not find a feasible coordination, the local subproblem is expanded.
In the worst case, the local subproblem is expanded until it is equivalent to the original problem. 
At this point, the completeness of the approach is dependent on the methods used in the planning hierarchy.
If the planning hierarchy includes a probabilistically complete method (e.g. Composite PRM) and the termination criteria is adjusted to allow it to continue to search for a path once the global problem setting is reached, then the approach is probabilistically complete.

ARC has no optimal guarantee, even if an asymptotically optimal method is used in the hierarchy.
This follows from the local resolution of every conflict.
If there exists a better resolution for the conflict outside the local subproblem, no method in the hierarchy will find it.

\section{Experiments}

In our experiments, we demonstrate that, when confronted with challenges involving numerous robots necessitating tight coordination, ARC stands out as the appropriate method to use, showcasing its capacity to dynamically adapt robot coordination. Furthermore, we show that, for scenarios demanding minimal coordination, ARC exhibits scalability comparable to that of a purely decoupled technique. Additionally, in situations mandating precise coordination, ARC performs with the same efficacy as a purely coupled technique.

\subsection{Experimental setup}

We evaluate the performance of our method across three distinct scenarios, each demanding low, high, and varying levels of robot coordination. For each category, we provide two scenarios: one involving mobile robots and the other featuring manipulator robots. We compare against decoupled (Decoupled PRM~\cite{sl-uppccdpmrs-2002}), hybrid (CBS-MP~\cite{smsm-romrmpucbs-21}), and coupled (Composite PRM~\cite{sl-uppccdpmrs-2002}) baselines. We ran 33 random trials for each scenario. Each trial was allotted 1000 seconds for planning,  after which the trial was considered a failure. We report the planning time to find the first solution and the cost of that solution for each method. The solution cost is calculated as the sum of costs, which is the total of all path timesteps, including waiting time.

\subsection{Scenarios}

    \begin{figure}[h!]
		\centering
		\begin{subfigure}[b]{0.30\linewidth}
			\includegraphics[width=\linewidth]{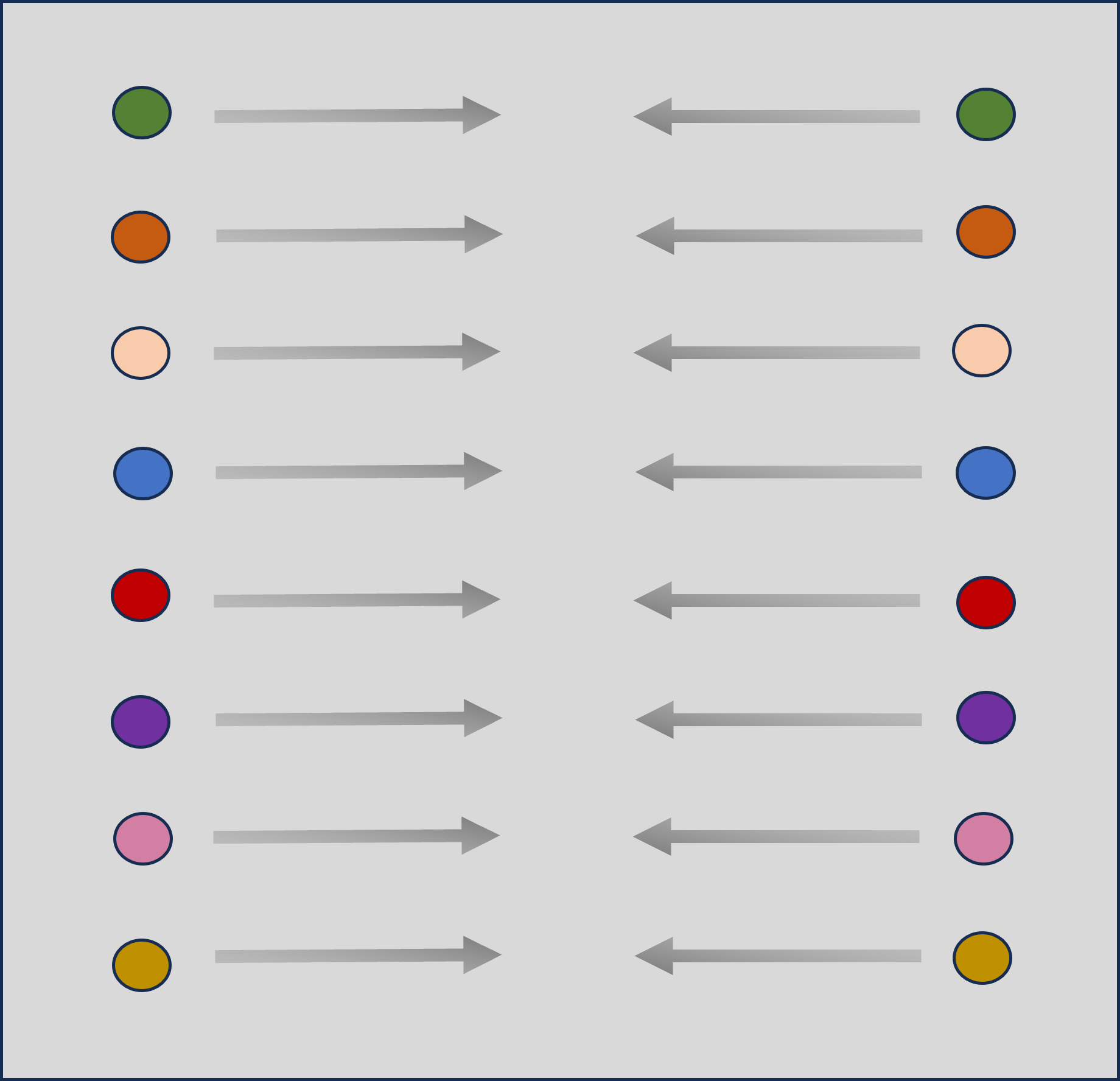}
			\caption{}
            \label{fig:LowCoordination-a}
		\end{subfigure}
		\begin{subfigure}[b]{0.30\linewidth}
			\includegraphics[width=\linewidth]{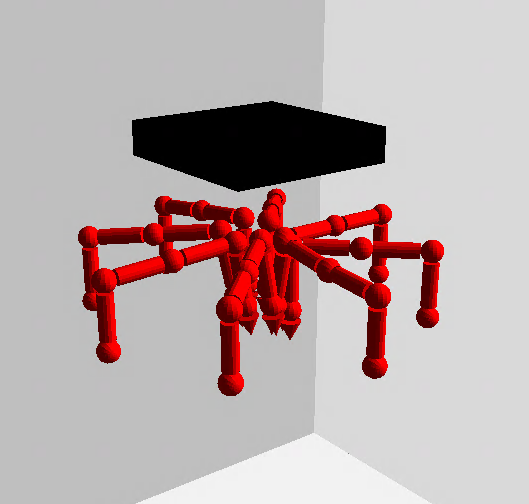}
			\caption{}
            \label{fig:LowCoordination-b}
		\end{subfigure}
		\begin{subfigure}[b]{0.31\linewidth}
			\includegraphics[width=\linewidth]{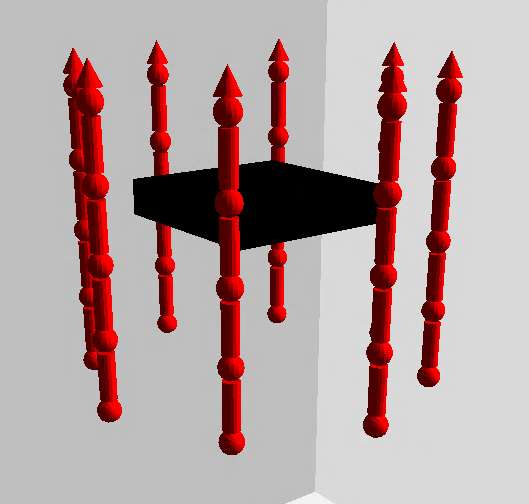}
			\caption{}
            \label{fig:LowCoordination-c}
		\end{subfigure}
		\caption{\small The low coordination scenarios. a) Each pair of robots on the same row must switch positions. The figure is scaled up for visibility. b) The manipulators' start configurations. c) The manipulators' goal configurations. \vspace{-2.5mm}}
		\label{fig:LowCoordination}
	\end{figure} 

    \begin{figure}[h!]
		\centering
		\begin{subfigure}[b]{0.39\linewidth}
			\includegraphics[width=\linewidth]{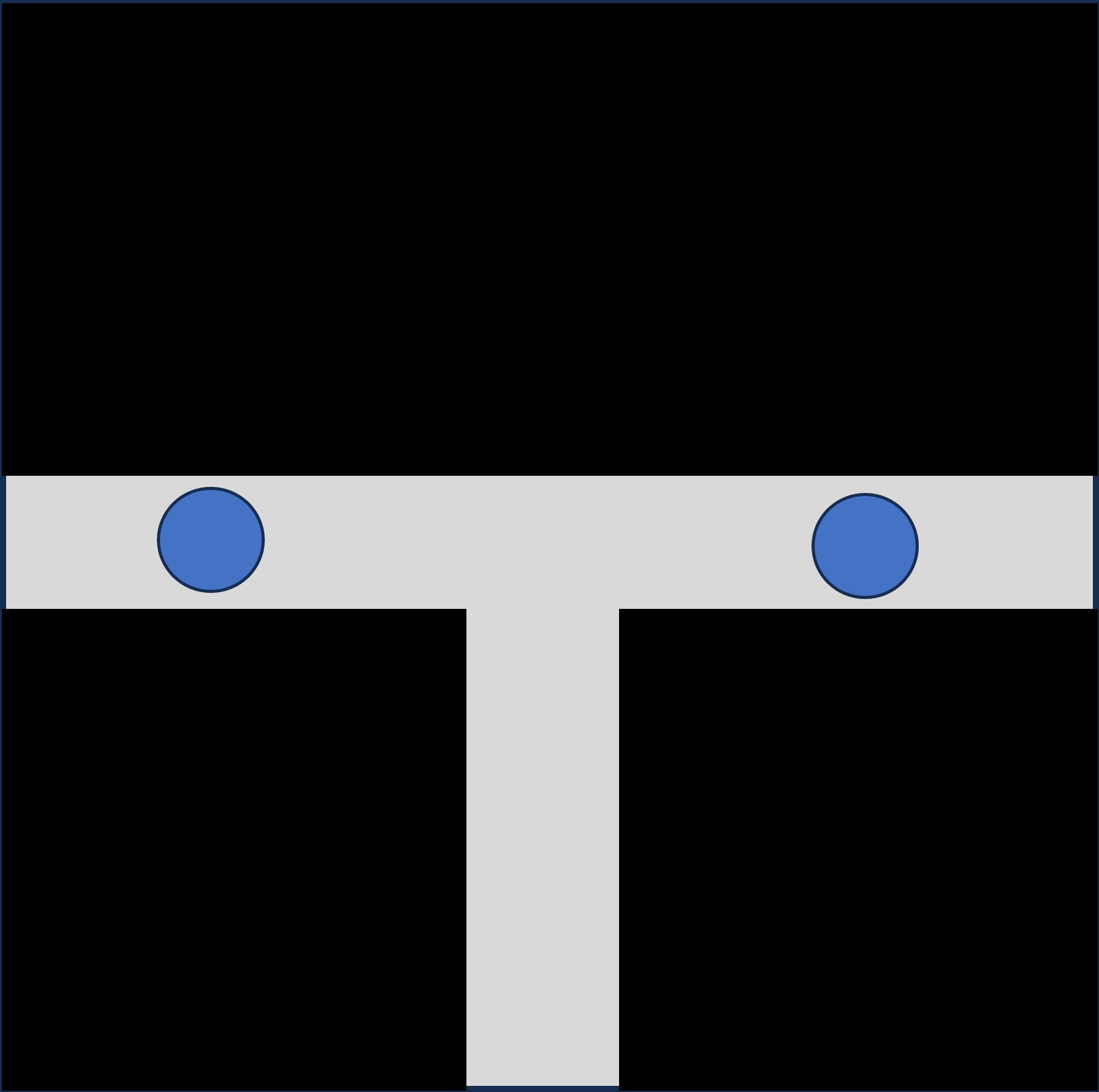}
			\caption{}
            \label{fig:HighCoordination-a}
		\end{subfigure}
		\begin{subfigure}[b]{0.39\linewidth}
			\includegraphics[width=\linewidth]{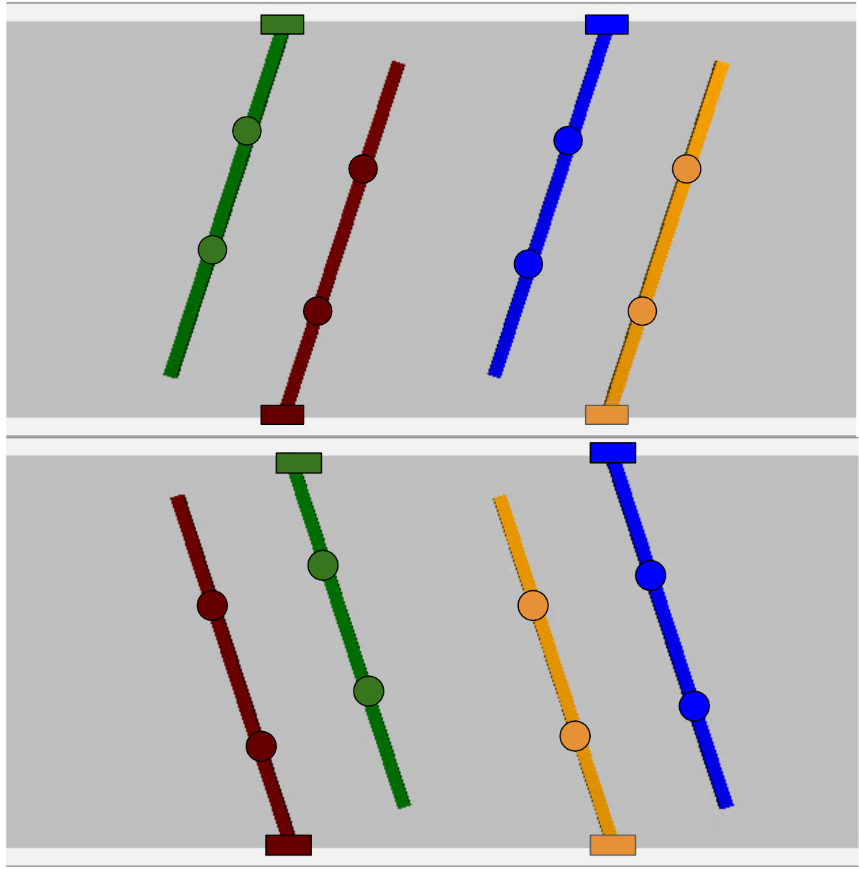}
			\caption{}
            \label{fig:HighCoordination-b}
		\end{subfigure} 
		\caption{\small The high coordination scenarios.  a) The inlet scenario where the two robots need to switch positions. The figure is scaled up for visibility. b) Each pair of manipulator arms needs to swing from left to right/right to left to reach their goal positions which requires one to move out of the way first. \vspace{-2mm}}
		\label{fig:HighCoordination}
	\end{figure}

\subsubsection{Scenario I: Low coordination} \label{subsec:scenario-2}
In this scenario, we examine instances where minimal or no coodination is needed. ARC proves competitive against decoupled approaches, showcasing effective scalability in these contexts.

For mobile robots, we examine the scenario illustrated in Fig. \ref{fig:LowCoordination-a}. Robots must switch positions horizontally, leading to numerous conflicts that do not demand high coordination. 
The problem is scaled by doubling the number of robots on each side.

For manipulator robots, we address the challenge of untangling the robots from a start (\ref{fig:LowCoordination-b}) to a goal position (\ref{fig:LowCoordination-c}). 
The problem is scaled by doubling the number of manipulators in a ring pattern.


\subsubsection{Scenario II: High coordination} \label{subsec:scenario-3}
In this scenario, we investigate situations requiring higher levels of coordination. We showcase that ARC provides the necessary coordination and competes effectively against a pure coupled approach, which excels in these scenarios.

For mobile robots, we examine a scenario with two robots swapping positions in a narrow passage with a central inlet (Fig. \ref{fig:HighCoordination-a}). Only one robot can pass the passage at a time, thus a solution demands precise coordination.

For manipulator robots, we examine 3-dof planar manipulators positioned oppositely (Fig. \ref{fig:HighCoordination-b}), with the top manipulators moving right and the bottom ones moving left.
More precise coordination is required, as robots need to find ways to contract themselves, creating enough space to avoid collisions.
We consider scenarios with both 2 and 4 manipulators.


\subsubsection{Scenario III: Adaptive coordination} \label{subsec:scenario-1}

In real-world applications, the required coordination between robots will likely be unknown. This scenario depicts realistic conditions where different coordination levels are needed at different stages of the problem. 

\begin{figure}[h!]
		\centering
		\begin{subfigure}[b]{0.32\linewidth}
			\includegraphics[width=\linewidth]{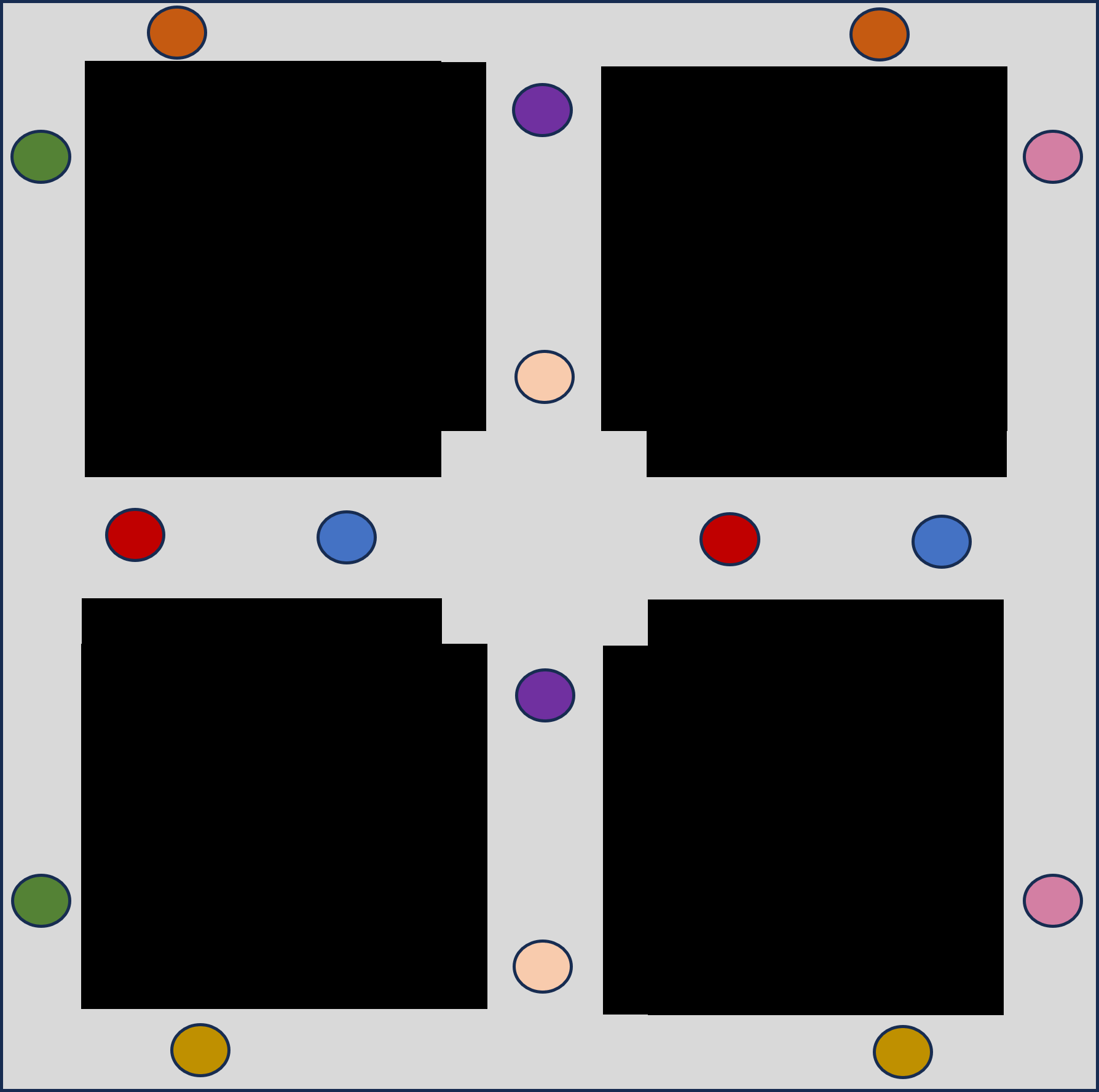}
			\caption{}
            \label{fig:AdaptiveCoordination-a}
		\end{subfigure}
		\begin{subfigure}[b]{0.32\linewidth}
			\includegraphics[width=\linewidth]{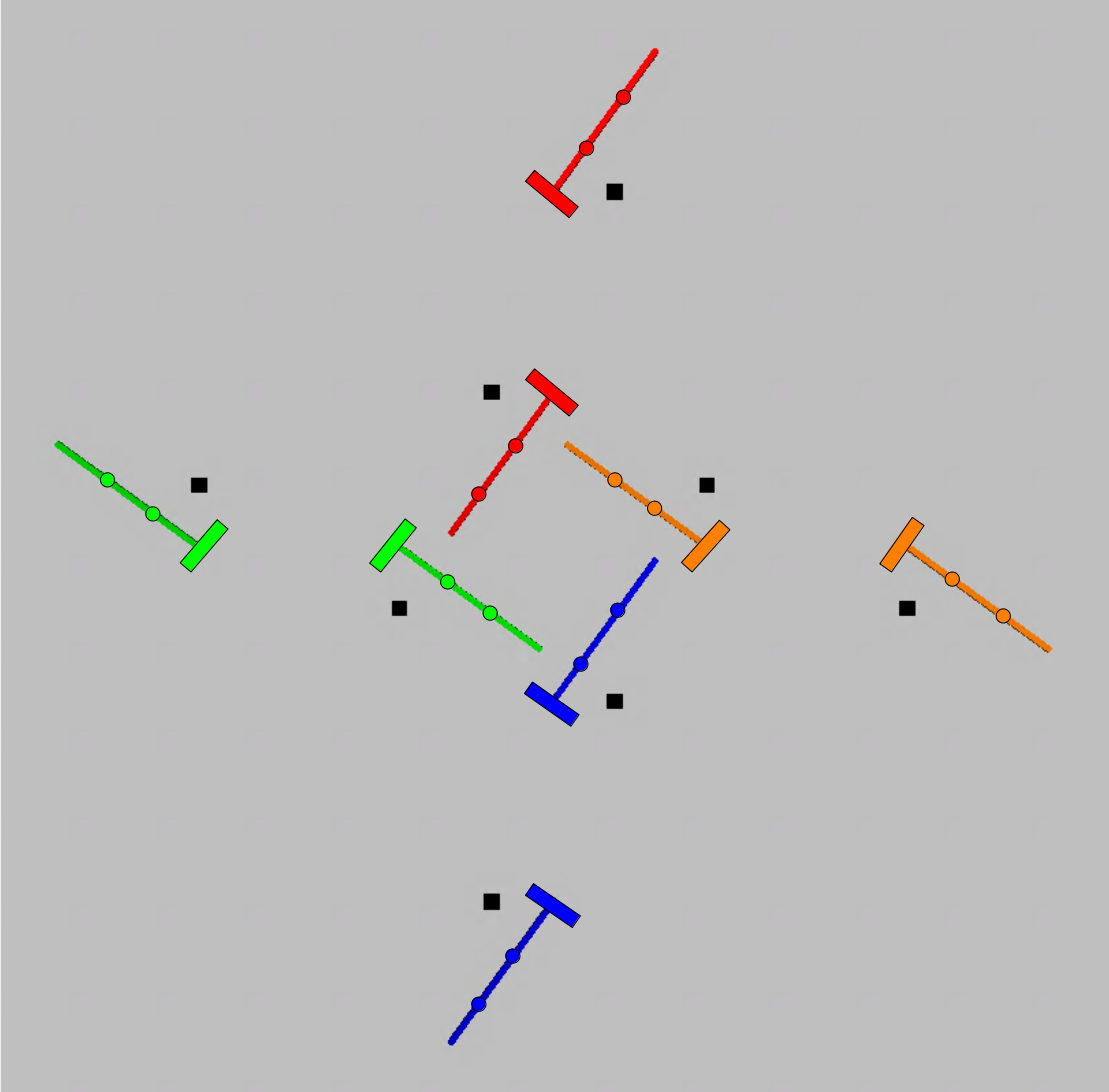}
			\caption{}
            \label{fig:AdaptiveCoordination-b}
		\end{subfigure}
		\begin{subfigure}[b]{0.32\linewidth}
			\includegraphics[width=\linewidth]{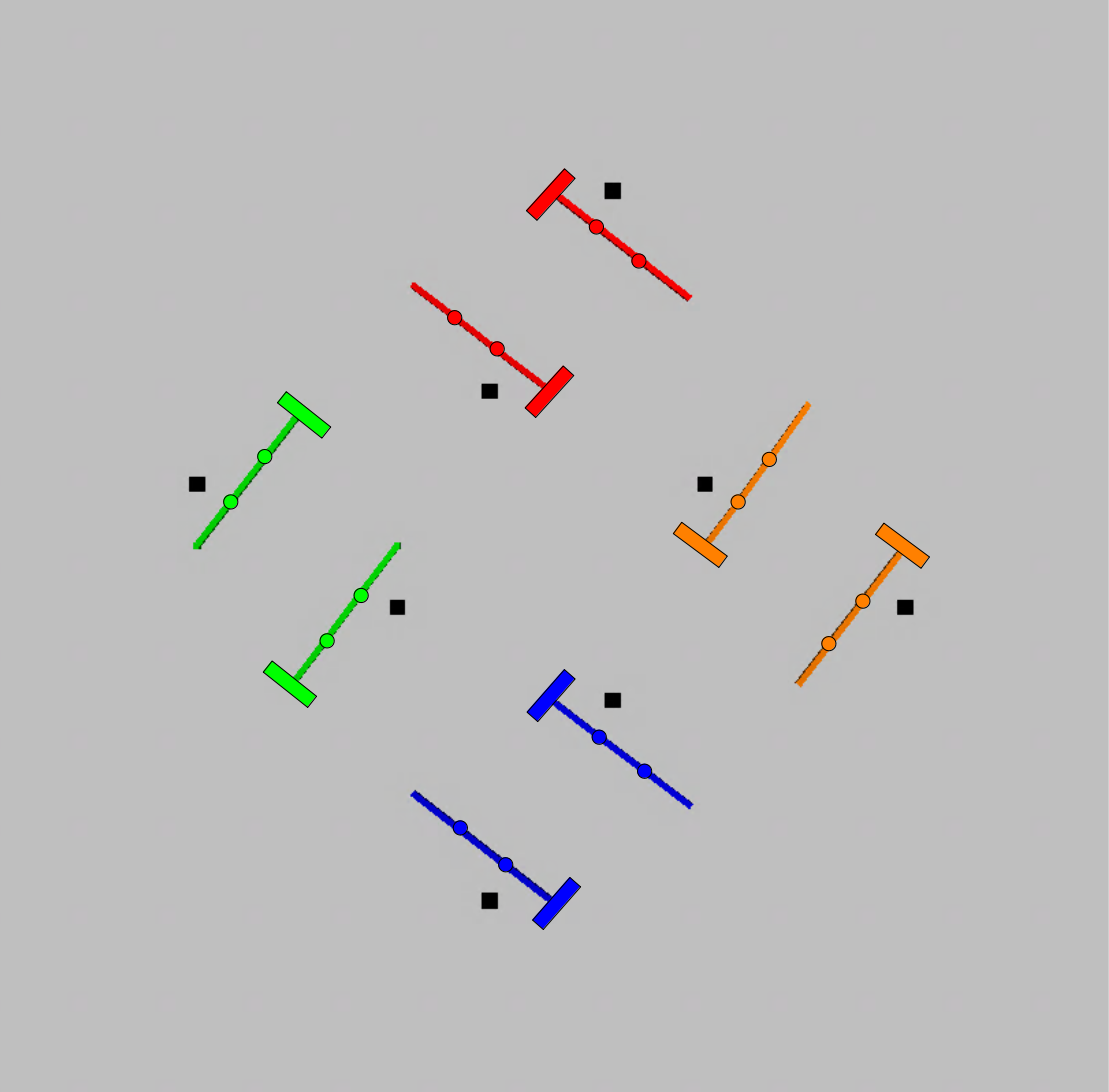}
			\caption{}
            \label{fig:AdaptiveCoordination-c}
		\end{subfigure}  
		\caption{\small The adaptive coordination scenarios. In both scenarios, conflicts involving varying numbers of robots are likely to occur. a) Mobile robots of the same colors must switch positions. The figure is scaled up for visibility. b,c) Start and goal configurations (the blocks at the end represent the bases) for the manipulator scenario.   
        \vspace{-5mm}
        }
		\label{fig:AdaptiveCoordination}
	\end{figure}

In the case of mobile robots, we examine 16 robots in a warehouse with narrow passages (\ref{fig:AdaptiveCoordination-a}). The number of robots involved in conflicts changes, necessitating varying levels of coordination. Two-robot conflicts typically arise along passages, while four-robot conflicts are more likely to occur in the center.

Regarding manipulator robots, we examine a scenario involving eight 3-dof planar manipulators with starting and target poses depicted in \ref{fig:AdaptiveCoordination-b} and \ref{fig:AdaptiveCoordination-c}. Initially, the four central manipulators encounter a 2-robot conflict with those on the outer rim. As the inner manipulators rotate towards the center, all four become involved in a conflict that requires higher coordination for resolution.


     \begin{table}[h!]
   \huge 
    \resizebox{\columnwidth}{!}{%
    
    \begin{tabular}{|c|c|c|cr|cr|c|}
    \hline
    \multirow{2}{*}{\textbf{\begin{tabular}[c]{@{}c@{}}Robot \\ type\end{tabular}}} & \multirow{2}{*}{\textbf{\begin{tabular}[c]{@{}c@{}}Number \\ of robots\end{tabular}}} & \multirow{2}{*}{\textbf{Method}} & \multicolumn{2}{c|}{\textbf{Planning Time}} & \multicolumn{2}{c|}{\textbf{Solution Cost}} & \multirow{2}{*}{\textbf{\begin{tabular}[c]{@{}c@{}}Success\\  rate\end{tabular}}} \\ \cline{4-7}
     &  &  & \multicolumn{1}{c|}{\textbf{Avg}} & \multicolumn{1}{c|}{\textbf{Std dev}} & \multicolumn{1}{c|}{\textbf{Avg}} & \multicolumn{1}{c|}{\textbf{Std dev}} &  \\ \hline
    Mobile & 16 & ARC & \multicolumn{1}{r|}{24.5} & 20.5 & \multicolumn{1}{r|}{450.1} & 19.6 & \multicolumn{1}{r|}{100\%} \\ \hline
    Manipulator & 8 & ARC & \multicolumn{1}{r|}{9.4} & 6.1 & \multicolumn{1}{r|}{316.5} & 121.5 & \multicolumn{1}{r|}{100\%} \\ \hline
    \end{tabular}%
    }
    \caption{\small{Results for the Adaptive Coordination Mobile and Manipulator scenarios}\vspace{-4mm} }
    \label{tab:adaptive-results}
    \end{table}

\subsection{Results}

\subsubsection{Scenario I: Low coordination} \label{subsec:results-scenario-2}

    \begin{figure*}[h!]
		\centering
		\begin{subfigure}[b]{.8\linewidth}
			\includegraphics[width=\linewidth]{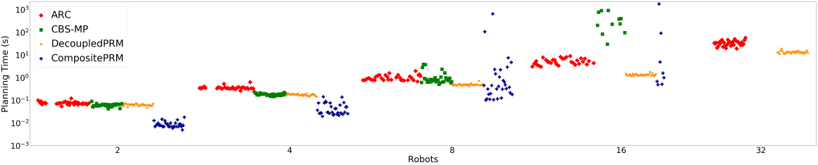}
		\end{subfigure}
		\begin{subfigure}[b]{.8\linewidth}
			\includegraphics[width=\linewidth]{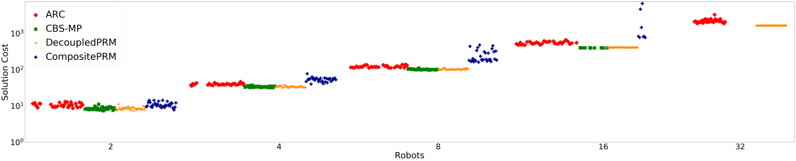}
		\end{subfigure}
		\caption{\small Results for Scenario II: Low Coordination Mobile Robots \vspace{-2mm}}
		\label{fig:low-mobile-results}
	\end{figure*} 

     \begin{figure*}[h!]
		\centering
		\begin{subfigure}[b]{.49\linewidth}
			\includegraphics[width=\linewidth]{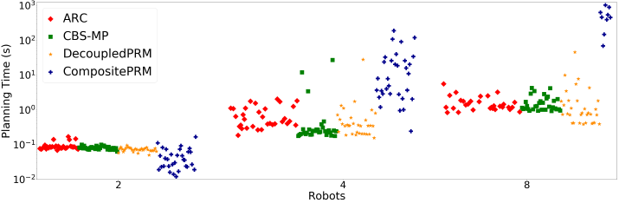}
		\end{subfigure}
		\begin{subfigure}[b]{.49\linewidth}
			\includegraphics[width=\linewidth]{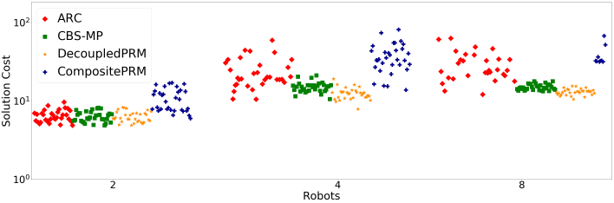}
		\end{subfigure}
		\caption{\small Results for Scenario II: Low Coordination Manipulator Robots\vspace{-2mm}}
		\label{fig:low-manipulator-results}
	\end{figure*} 
 For mobile robots, ARC and Decoupled PRM exhibit improved scalability as they are the only methods able to solve all the trials for 32 robots (Fig. \ref{fig:low-mobile-results}).  This is due to ARC behaving decoupled most of the time. Besides, resolving conflicts requires little effort. Given its thorough search for optimal solutions, CBS-MP successfully completed all trials for 8 robots but only achieved a 36\% success rate for trials involving 16 robots. Likewise, due to its inherently coupled behavior, Composite PRM can successfully plan for all trials involving 8 robots but achieves only a 24\% success rate for trials with 16 robots. 
 
Regarding manipulators, ARC and CBS-MP exhibit the best planning times, as are the methods that can handle better coordination and scalability (Fig. \ref{fig:low-manipulator-results}).  CBS-MP produces better solution costs due to its optimality, while ARC's solutions, though slightly more expensive, remain competitive. DecoupledPRM can also solve all the trials but at a slower pace, as it requires the exploration of more decoupled states. CompositePRM can only solve a 27\% of the  trials for eight robots due to its coupled behavior. 

\subsubsection{Scenario II: High coordination} \label{subsec:results-scenario-3}

    \begin{figure}[h!]
		\centering
		\begin{subfigure}[b]{0.34\linewidth}
			\includegraphics[width=\linewidth]{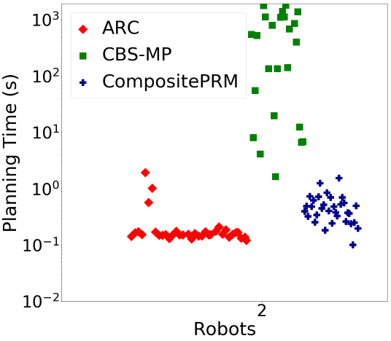}
		\end{subfigure}
		\begin{subfigure}[b]{0.62\linewidth}
			\includegraphics[width=\linewidth]{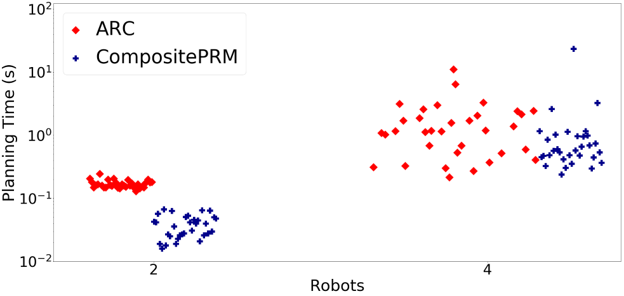}
		\end{subfigure}
		\begin{subfigure}[b]{0.34\linewidth}
			\includegraphics[width=\linewidth]{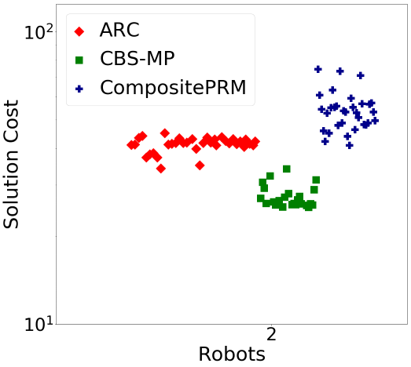}
			\caption{}
            \label{fig:high-mobile-results}
		\end{subfigure}
		\begin{subfigure}[b]{0.62\linewidth}
			\includegraphics[width=\linewidth]{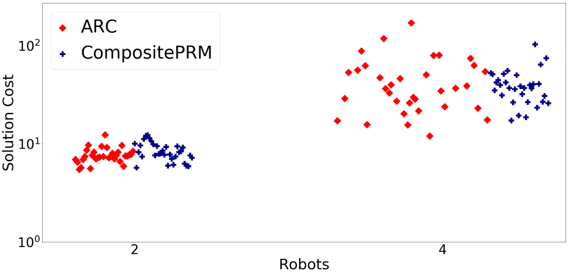}
			\caption{}
            \label{fig:high-manipulator-results}
		\end{subfigure}
		\caption{\small Results for Scenario III: a) High Coordination Mobile, b) High Coordination Manipulator \vspace{-1mm}}
		\label{fig:high-results}
	\end{figure}

 For mobile robots, ARC and CompositePRM are the only methods able to solve all the trials (Fig. \ref{fig:high-mobile-results}). They can directly explore the composite space, allowing for encountering the necessary coupled transitions to solve the problem. However, ARC produces better cost solutions by concentrating composite exploration in the inlet region. CBS-MP solved only 66\% of the trials because, in some cases, the decoupled roadmaps lack the required states for coordinating the robots at the inlet. DecoupledPRM failed to solve all the trials due to its lack of coordination.

Regarding  manipulators, ARC and CompositePRM are the only methods able capable of solving the trials, showcasing comparable performance in both planning time and cost solutions. ARC has slightly higher planning times, attributed to resolving conflicts before integrating the four robots, while CompositePRM plans them all simultaneously. On the other hand, CBS-MP falls short as it exhausts time in an extensive search of the decoupled roadmaps, and DecoupledPRM encounters fails due to its inherent lack of coordination.

\subsubsection{Scenario III: Adaptive coordination} \label{subsec:results-scenario-1}
In both scenarios, (Table \ref{tab:adaptive-results}) we observe that only ARC can generate feasible solutions.
This occurs because ARC is the only method able to dynamically adapt robot coordination to resolve the diverse conflicts that arise along the paths, where each conflict requires different levels of coordination.
In contrast, the other methods fail due to either coordination deficiencies or their incapacity to scale effectively for a larger number of robots.
In Fig. \ref{fig:path-composition}, we illustrate a trial and how the ARC final paths of Figs. \ref{fig:AdaptiveCoordination-b}-\ref{fig:AdaptiveCoordination-c} result from transitioning through path segments planned in state spaces of different dimensionalities.

In mobile robot scenarios, ARC's planning time varies due to conflicts requiring high coordination. Using a sampling-based method for conflict resolution introduces randomness, leading to increasing variability as more conflicts are addressed.
In manipulator robot scenarios, ARC's solution costs also vary because ARC stops local exploration when the conflict resolution is found. Due to the complexity of manipulator planning, achieving lower-cost conflict resolutions requires a deeper exploration of the planning space, inevitably increasing planning time, a factor not within the focus of this work.

	\section{Conclusions and Future Work}


    In this paper, we introduce ARC, a novel hybrid MRMP method that employs a subproblem-based approach for resolving robot conflicts. ARC efficiently explores the extensive multi-robot planning space by introducing local subproblems, whichs enable a cost-effective exploration of relevant regions within the composite space. The solutions to these subproblems depict the appropriate robot motions for conflict resolution. ARC can rapidly adapt subproblems to plan for the necessary robots and physical space.

The results presented in this paper demonstrate ARC's ability to offer simultaneous scalability and coordination across various scenarios. In comparison to the decoupled baseline, ARC exhibits competitive scalability, and it also competes effectively in terms of coordination against the coupled approach. In the scenarios featuring a large number of robots with varying coordination requirements, ARC stands out as the only method capable of finding solutions, thanks to its capacity to quickly adapt subproblems for the resolution of diverse type of conflicts.

Looking ahead, our future research will focus on exploring additional heuristics to enhance subproblem adaptation for improved conflict resolution. We will also investigate efficient approaches to reuse local solutions when resolving similar conflicts, thus reducing the need for re-planning.

	
	\bibliographystyle{ieeetr}
	\bibliography{robotics.bib}

\end{document}